\pdfoutput=1

\documentclass[11pt]{article}

\usepackage{acl}

\usepackage{times}
\usepackage{latexsym}
\usepackage{graphicx}
\usepackage{times}
\usepackage{latexsym}
\usepackage{latexsym}
\usepackage{amssymb}
\usepackage{booktabs}
\usepackage{amsmath}
\usepackage{booktabs}
\usepackage{subfig}
\usepackage{graphicx}
\usepackage{xspace}
\usepackage{times}
\usepackage{latexsym}
\usepackage{multirow}
\definecolor{myblue}{rgb}{0.9, 0.1, 0.94}
\definecolor{mygreen}{rgb}{0.64, 0.56, 0.88}
\definecolor{myyellow}{rgb}{0.68, 0.64, 0.35}
\definecolor{mygreen}{rgb}{0.68, 0.9, 0.6}

\newcommand{\gn}[1]{\textcolor{mygreen}{\bf\small [#1 --gn]}}

\usepackage{microtype}

\title{Searching for Effective Tuning Strategies for\\ Multilingual Summarization}
\title{Searching for Effective Multilingual Fine-Tuning Methods: \\ A Case Study in Summarization}

 \author{Yiwei Qin$^\spadesuit$ \;\; Graham Neubig$^{\spadesuit\clubsuit}$\;\;  Pengfei Liu$^{\spadesuit\clubsuit}$\thanks{\ \ Corresponding author} \\ 
 \\
  $^{\spadesuit}$Carnegie Mellon University, $^{\clubsuit}$Inspired Cognition\\ 
  \texttt{\{yiweiq,gneubig,pliu3\}@cs.cmu.edu} \\
  }

\date{}

\begin{document}
\maketitle
\begin{abstract}

Recently, a large number of tuning strategies have been proposed to adapt pre-trained language models to downstream tasks. In this paper, we perform an extensive empirical evaluation of various tuning strategies for multilingual learning, particularly in the context of text summarization.
Specifically, we explore the relative advantages of three families of multilingual tuning strategies (a total of five models) and empirically evaluate them for summarization over 45 languages.  Experimentally, we not only established a new state-of-the-art on the \texttt{XL-Sum} dataset but also derive a series of observations that hopefully can provide hints for future research on the design of multilingual tuning strategies.%
\footnote{
Code at \url{https://github.com/qinyiwei/Multi-Sum.git}
}
\end{abstract}

\section{Introduction} \label{Introduction}








Methods that perform fine-tuning of pre-trained language models (PLMs) now represent the state-of-the-art across a wide variety of NLP tasks~\cite{howard2018universal,han2021pre}. 
Because there are a myriad of methods for tackling this important task of fine-tuning LMs, there is an increasing body of research investigating the empirical strengths and weaknesses of different tuning strategies across several tasks 
\cite{peters-etal-2019-tune,mahabadi2021compacter,karimi-mahabadi-etal-2021-parameter,li-liang-2021-prefix,mao2021unipelt,hu2021lora,min2021noisy,he2021towards}.
One of the major design dimensions of these works revolves around
\textit{which set of model parameters are updated}; should fine-tuning only adjust a few additional parameters that are not part of the initial LMs (e.g., Adapters \citep{houlsby2019parameter}, or Prefix Tuning \citep{li-liang-2021-prefix,xue-etal-2021-mt5}),   update all parameters of the pre-trained models~\citep{dai2015semi,BERT,schick-schutze-2021-shot,fu2022polyglot}, or update only a subset of parameters~\citep{diff-pruning}?

At the same time, there has been much progress in multilingual models based on pre-trained LMs \cite{lample2019cross,conneau-etal-2020-unsupervised,liu-etal-2020-multilingual-denoising}.
However, there is a notable gap in the literature -- to our knowledge, there are no comprehensive comparative studies on how different tuning strategies behave in multi-lingual scenarios -- there is significant work on multilingual adapters \citep{pfeiffer-etal-2020-mad,ansell-etal-2021-mad-g} and parameter tying across languages~\cite{sachan-neubig-2018-parameter,lin-etal-2021-learning}, but few studies comparing different families of methods.

\begin{figure}[t]
\centering
\small
\subfloat[ ]{{\includegraphics[width=0.25\linewidth]{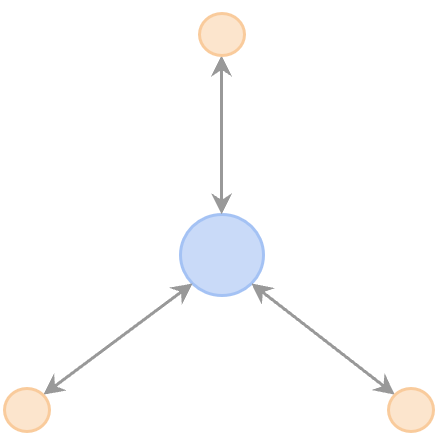} }}
\hspace{15pt}
\subfloat[\centering]{{\includegraphics[width=0.25\linewidth]{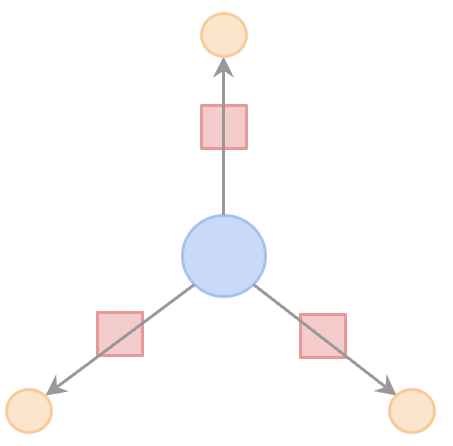} }}%
\hspace{15pt}
\subfloat[\centering ]{{\includegraphics[width=0.25\linewidth]{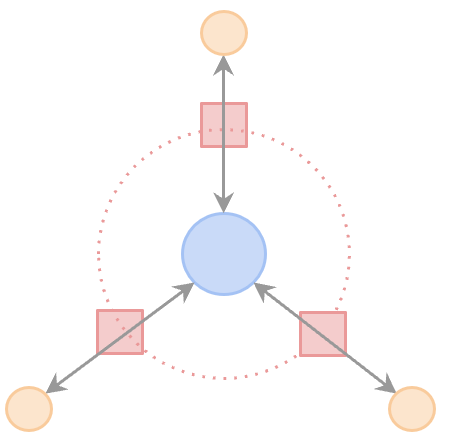} }}%
\caption{Different frameworks for multilingual learning, where orange circles represent different languages and blue circles denote pre-trained language models (PLMs). Red boxes refer to additional learnable parameters, such as adapters or prefixes. Double sided arrows represent that the parameters of PLMs are tunable.
}%
\label{fig:intro} 
\end{figure}

In this paper, we try to fill this gap by performing a comprehensive study of different parameter tuning techniques in the context of text summarization~\cite{rush-etal-2015-neural, nallapati-etal-2016-abstractive, chopra-etal-2016-abstractive, BART, zhang2020pegasus,dou-etal-2021-gsum}.
We focus particularly on summarization as previous work on parameter-efficient tuning has noted that the differences between tuning techniques are particularly salient in more complex generative tasks such as summarization, as opposed to text classification \citep{he2021towards}.
We draw on the techniques examined in the monolingual scenario and combine the unique characteristics of the multilingual scenario (e.g., shared features across languages) to derive different architectures for multilingual learning. These frameworks encompass some existing works on multilingual learning, but also allow us to propose new learning methods and perform comparisons between different frameworks.

\autoref{fig:intro} (a) shows a commonly-used framework~\cite{hasan-etal-2021-xl} in which one tunable pre-trained model is shared by different languages.
\autoref{fig:intro} (b) introduces a language-specific module with learnable parameters while keeping the PLM's 
parameter frozen, i.e.~parameter-efficient tuning \cite{mao2021unipelt,he2021towards}. In practice, the language-specific module could be instantiated as an adapter, prefix, or other variety of extra parameters.
Notably, this kind of language-specific module is independent for each language and cannot share information, thus low-resource languages cannot benefit from other related languages.
\autoref{fig:intro} (c) tries to alleviate this problem in two ways: making parameters of pre-trained models tunable or introducing additional modules whose parameters can be shared by different languages.


Using this framework, we ask these questions:


Q1: How well do different parameter-efficient tuning methods (\autoref{fig:intro}-b) perform compared to PLM fine-tuning models (\autoref{fig:intro}-a) in multi-lingual summarization?
Q2: Will supervised transfer, a commonly used technique in multi-lingual learning, be helpful for parameter-efficient tuning?
Q3: Could better results be achieved by enabling information exchange between different languages? How do different choices of parameter-efficient tuning methods interact with this sharing?

We explore these questions by performing extensive experiments over 45 different languages
.
Our quantitative and qualitative analyses find several observations, such as:\\
(1) Compared to PLM fine-tuning, both parameter-efficient tuning methods (prefix- and adapter-tuning) are advantageous in low-resource scenarios. 
Particularly, prefix-tuning outperforms adapter-tuning with extremely few samples over different languages \S\ref{exp1}.
(2) Parameter-efficient tuning is possible to fail in the \textit{supervised transfer} setting (\S\ref{exp2}), where pre-trained language models are fine-tuned on the source languages whose scripts are distant from the target language's.
(3) Adding language specific adapters or prefixes 
while additionally tuning the PLM's parameters,
 can maintain multi-lingual PLM fine-tuning’s advantage of sharing information among languages, as well as preserving private parameters for each language to reduce the negative effect of the limited capacity of one LM shared by all languages. We achieve a new state-of-the-art performance with such a multi-lingual tuning strategy.

\section{Preliminaries}

\subsection{Task Formulation}
Abstractive summarization can be formulated as a conditional generation task where the input ${D}$ is a document, and the output ${S}$ is a short summary. The majority of state-of-the-art models for abstractive summarization use encoder-decoder models \cite{sutskever2014sequence}, where an encoder generates representations for the source document $\mathbf{D} = [\mathbf{d}_1,...,\mathbf{d}_{m}]$, and a decoder outputs the summary $\mathbf{S} = [\mathbf{s}_1,...,\mathbf{s}_{n}]$ one target token at a time. The conditional probability of a single sample is modeled as $p(\mathbf{s}^i|\mathbf{d}^i;\theta)$,  
and hence parameters $\theta$ are obtained by maximum likelihood estimation 
\begin{equation}
    \mathop{\mathrm{argmax}}\limits_\theta \sum_{(\mathbf{d}^i,\mathbf{s}^i)\in(D,S)}\log p(\mathbf{s}^i|\mathbf{d}^i;\theta),
\end{equation}
where $(D,S)$ is the parallel training corpus. For multilingual text summarization, $D$ and $S$ can be in any of a number of languages.



\subsection{Tuning Strategy}
Recently, applying pre-trained language models (PLMs) to  abstractive summarization tasks equipped with diverse tuning strategies has achieved a great success, which can be formulated as below: 
\begin{align}
    h_i = \textsc{PLM}(D, s_i, h_{<i};\theta_{\text{plm}},\theta_{\text{add}})
\end{align}
where $\text{PLM}$ is a sequence to sequence pre-traiend LMs (e.g., T5~\citep{T5} or BART~\citep{BART}), $\theta_{\text{plm}}$ represents the original PLM parameter and $\theta_{\text{add}}$ denotes the additional parameters added by different tuning strategies.

\begin{figure*}[!t]
\centering
\subfloat[\centering PLM Fine-tuning]{{\includegraphics[width=3.5cm]{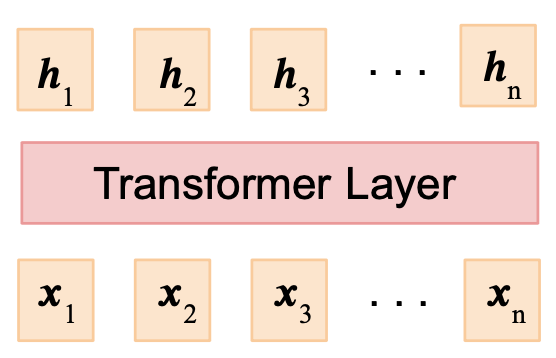} }}%
\hspace{10mm}
\subfloat[\centering Adapter-tuning]{{\includegraphics[width=3.5cm]{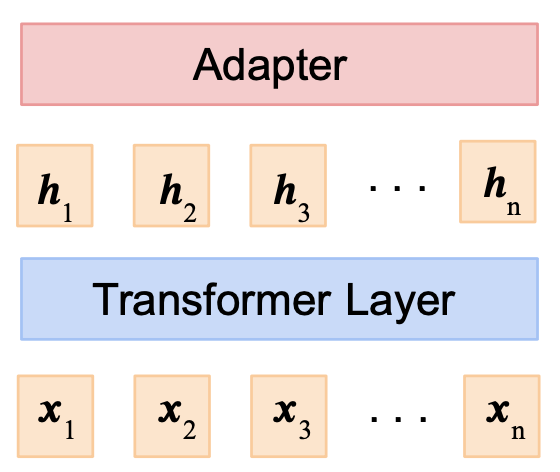} }}%
\hspace{10mm}
\subfloat[\centering Prefix-tuning]{{\includegraphics[width=3.5cm]{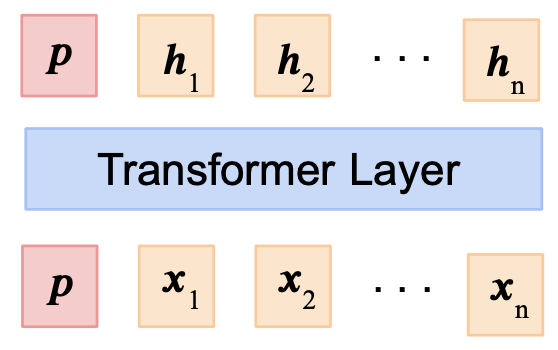} }}%
\caption{Different tuning methods. Red indicates trainable parameters; blue indicates frozen parameters.}%
\label{tuning_methods} 
\end{figure*}

Based on whether and when parameters $\theta_{\text{plm}}$ and $\theta_{\text{add}}$ will be tuned, different tuning strategies as illustrated in Fig.~\ref{tuning_methods} have been explored, which we will detail below for the better introduction of multi-lingual tuning strategies.


\paragraph{PLM Fine-tuning}
This is one of the most common tuning strategies that aim to tune all of the parameters $\theta_{\text{plm}}$.
While PLM fine-tuning has achieved strong performance on many benchmarks, one major limitation lies in the requirement of large training samples, which is not feasible in the low-resource scenario.

To alleviate this issue, parameter-efficient tuning has been extensively explored recently,  among which we select two representative methods which are initially designed for generation tasks, consistent with our goal.

\paragraph{Adapter-tuning}

Adapter-tuning adds additional lightweight layers between the layers of an existing PLM. Although there is a variety of ways to define ``adapter'', we adopt the definition of \cite{bapna2019simple}.
Specifically, the adapter block consists of 
(1) a layer normalization $\mathrm{LN}(\cdot)$ for the input of the adapters, 
(2) an autoencoder whose inner dimension can be adjusted according to the complexity of the target task with a down projection layer, an up projection layer, and a nonlinear activation function between them, and 
(3) a residual connection. 
Formally, given $h_i \in \mathbb{R}^{d}$ be the output of i-th layer, the adapter is formulated: 
\begin{equation*}
    \textsc{Adapter}(h_i)=(\mathrm{ReLU}(\mathrm{LN}(h_i)\mathbf{W}_i^{db}))\mathbf{W}_i^{bd}+h_i,
\end{equation*}
where  b is the inner dimension, $\mathbf{W}_i^{db} \in \mathbb{R}^{d\times b}$ is the weight of down projection layer and $\mathbf{W}_i^{bd} \in \mathbb{R}^{b\times d}$ is the weight of up projection layer.

\paragraph{Prefix-tuning }
Prefix-tuning~\cite{li-liang-2021-prefix} prepends a prefix for every layer of a LM. Let $\mathbf{H}_i^{LM} \in \mathbb{R}^{t \times d}$, where $d$ is the hidden dimension of LM, $t$ is the input sequence length, denote the hidden representation of the i-th layer. We prepend prefixes at each layer to obtain $\mathbf{H}_i=[\mathrm{Prefix}_i; \mathbf{H}_i^{LM}] \in \mathbb{R}^{(t+l) \times d}$, where $l$ is the prefix length, $\mathrm{Prefix}_i \in \mathbb{R}^{l \times d}$ is prepended prefix. 

We can look up trainable matrix $\mathbf{P}_\theta \in \mathbb{R}^{l \times (d \times n)}$, where $n$ is the number of layers of the LM, to get $\mathrm{Prefix}_i$. However, according to \cite{li-liang-2021-prefix}, reparameterization has better performance than directly updating $\mathbf{P}_\theta $ in practice. So we reparametrize the matrix $\mathbf{P}_\theta = \mathrm{MLP}(\mathbf{P}_\theta^{\prime})$, where $\mathrm{MLP}(\cdot)$ has the structure of an autoencoder with a tunable middle dimension size, and $\mathbf{P}_\theta^{\prime}$ is a smaller matrix with dimension $l \times d$.




\subsection{Multilingual Tuning Methods}

Based on the above-mentioned tuning strategies in single language scenarios, we investigate three different multilingual learning frameworks and explore their applicable scenarios in detail. 


\paragraph{Multilingual PLM Fine-tuning (MPF)}
This is a commonly-used setting~\citep{hasan-etal-2021-xl} when training samples from different languages are provided. Summarization systems share one multilingual pre-trained language model whose parameters can be updated by any system.

\paragraph{Multilingual Parameter-efficient Tuning (MPE) }
In this framework, additional private parameterized modules such as prefix or adapter are introduced for each system besides one shared multilingual pre-trained language model, whose parameters keep frozen. Some existing works~\cite{bapna2019simple} follow this framework but mainly focus on the use of adapters. 


\paragraph{Multilingual Private-shared Tuning (MPS)}
In the above method, although systems of different languages share one pre-training model, their parameters cannot be modified, which results in the lack of information interaction across languages and the difficulty in mining the shared knowledge.
In this framework, parameters from both additional modules and pre-trained models can be updated.

\section{Experiments}

\paragraph{Dataset}
As our evaluation testbed, we use the \texttt{XL-Sum} corpus \citep{hasan-etal-2021-xl},\footnote{License: CC BY-NC-SA 4.0.} which is a news dataset containing 1.1 million article-summary pairs in 45 languages. The dataset is collected from the British Broadcasting Corporation (BBC) website, using a bold paragraph at the beginning of each article
as the summary and the rest of the article as the input text. We choose XL-sum 
for its: (1) high language coverage, including low-resource, medium-resource, and high-resource languages, (2) similar intrinsic characteristics, e.g. novel n-gram ratio, abstractivity, and compression among all samples, allowing our analysis to focus on the differences across languages, other than different intrinsic features across samples. 

\paragraph{Evaluation Metric}
As is standard in summarization, we use ROUGE  \citep{rouge}  as our evaluation metric, which computes the n-gram similarity between the gold and the generated summary.\footnote{We use the rouge package provided by \texttt{XL-Sum} \citep{hasan-etal-2021-xl} to support multiple languages.}

\subsection{Exp-I: Effect of Tuning Method} \label{exp1}



To answer the question of how well different parameter-efficient tuning methods behave compared to standard LM fine-tuning in the multilingual setting (Q1), we study the performance of three tuning methods: prefix-tuning, adapter-tuning and PLM fine-tuning on different languages.

\subsubsection{Experiment Details}
\paragraph{Settings and Hyper-parameters} We use the base version of multilingual T5 \citep{xue-etal-2021-mt5} as a backbone, which covers most languages in \texttt{XL-Sum} dataset and is the same as \citep{hasan-etal-2021-xl}, allowing us to make a fair comparison. In our experiment, $\mathrm{MLP}$ of prefix-tuning is two linear layers with an inner dimension as a hyper-parameter. For prefix-tuning, the hyper-parameters we tune\footnote{Specifically, the number of epochs, batch size, learning rate, prefix length and inner dimension during training, beam search size and length 
penalty during inference.} are the same as \citet{li-liang-2021-prefix}. For adapter-tuning, the hyper-parameters remain the same as prefix-tuning except the prefix length that is not needed for adapter-tuning. More details are in the appendix. For both adapter-tuning and prefix-tuning, these hyper-parameters lead to about 8\% additional parameters compared to the LM’s total parameters, which are tuned during training while the LM’s parameters are frozen.
 To study whether language features will influence the choice of tuning methods, we choose five languages from different language families: \texttt{English} (Germanic), \texttt{Chinese simplified} (Sino Tibetan), \texttt{Spanish} (Romance)
, \texttt{Ukrainian} (Balto Slavic) and \texttt{Urdu} (Indo Iranian). 

We subsample the full dataset of each language to obtain sub-datasets of various sizes,\footnote{Concretely, \{5, 10, 20, 50, 100, 200, 500, 3000, 6000, 10000, 20000, 30000\}. For English, which has far more training samples than other languages, we add two training set size 100000 and 300000.} and sub-datasets of size $\le 500$ are considered ``few-shot'' experiments.  For each few-shot experiment, we randomly sample 3 different training sets and development set (with dev size = 20\% training set size). The reported result is the average of 3 experiments on the full test set of the chosen language.
The performance of few shot experiments is influenced by the training samples chosen \citep{zhao2021closer}, so we keep the sampled training set and development set the same for the three tuning methods to have a fair comparison. For non-few shot experiments, each size has one experiment and is tested on the full test set of the chosen language. The hyperparameters are chosen from a single language (Japanese) for each tuning method and applied as-is to all languages. 
We use the result from the checkpoint 
with the best validation set performance over all training epochs.

\begin{figure*}[t]
\centering
\subfloat[\centering English]    {\includegraphics[width=0.3\linewidth]{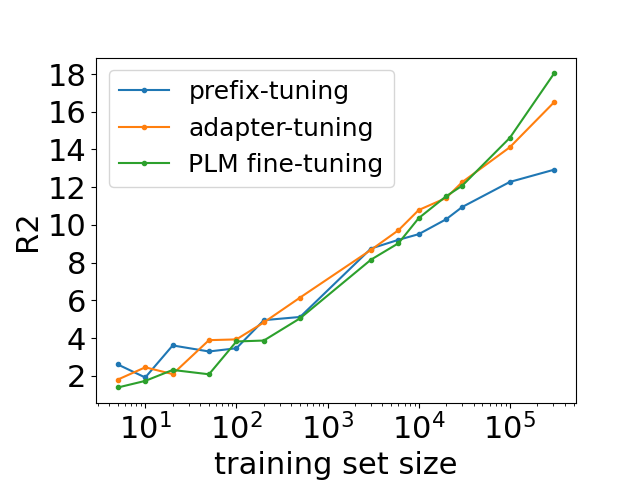}}
\hspace{10pt}
\subfloat[\centering Chinese Simplified]{{\includegraphics[width=0.3\linewidth]{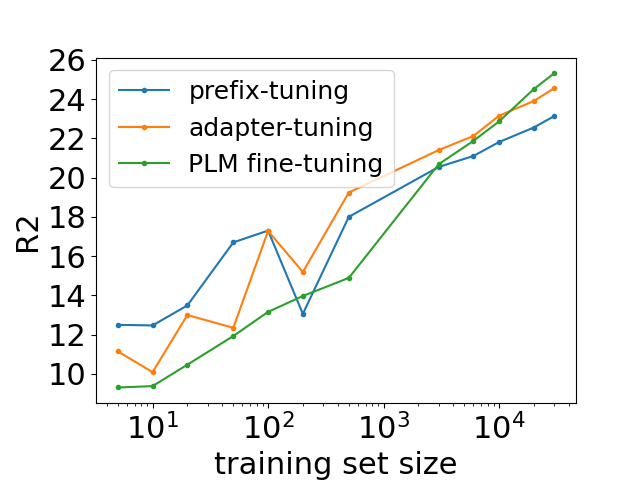} }}
\hspace{10pt}
\subfloat[\centering Spanish]{{\includegraphics[width=0.3\linewidth]{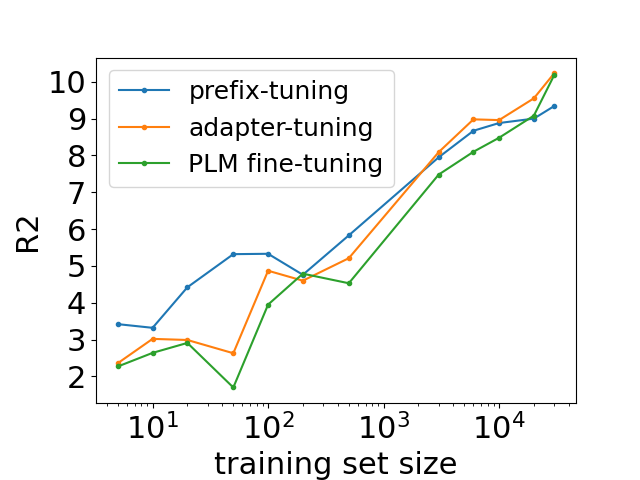} }}
\\
\subfloat[\centering Ukrainian]{{\includegraphics[width=0.3\linewidth]{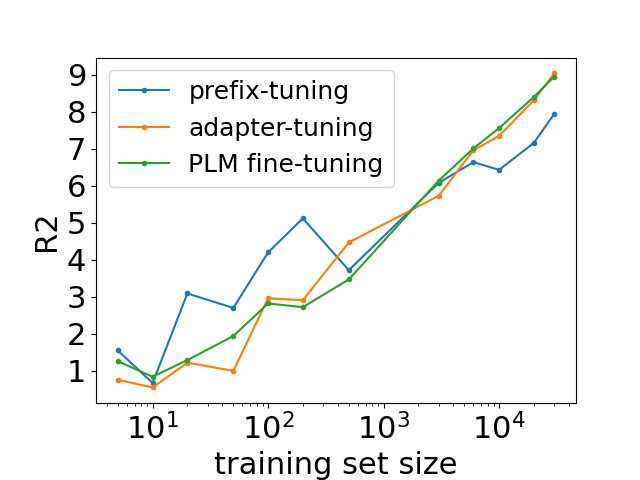} }}
\hspace{10pt}
\subfloat[\centering Urdu]{{\includegraphics[width=0.3\linewidth]{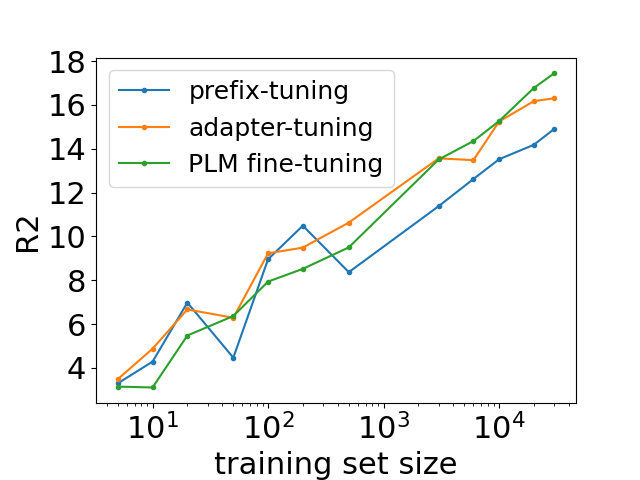} }}%
\hspace{10pt}
\subfloat[\centering Average]{{\includegraphics[width=0.3\linewidth]{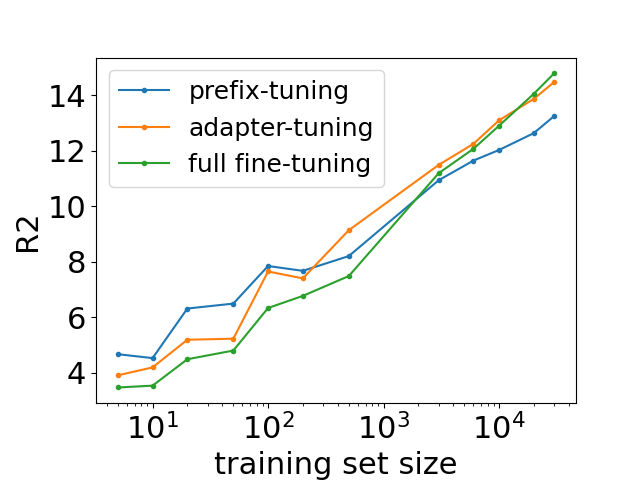} }}%
\caption{Performance of prefix-tuning, adapter-tuning and PLM fine-tuning on five languages over training set sizes. The x axis is the number of training samples at log scale, the y axis is the ROUGE-2 score.}%
\label{num_train_size} 
\end{figure*}

\subsubsection{Results and Analysis}

\paragraph{Results} Fig.\ref{num_train_size} illustrates the performance of three different tuning methods with respect to the available training samples, observations are: \\
(1) In general when the sample number is less than 200 prefix-tuning achieves the best performance. Between 200 and 10k adapter-tuning is superior, and greater than 10k PLM fine-tuning surpasses the other two. This indicates that \textit{regarding both the performance and parameter efficiency (only tuning 8\% of the parameters of PLM fine-tuning), prefix-tuning is the best choice when we have extremely few samples, while adapter-tuning is the winner in medium-resource settings}. 
(2) As the training set size increases from few shot to high-resource, PLM fine-tuning has the largest performance improvement, while prefix-tuning has the least performance improvement and adapter-tuning is the middle. 
(3) Compared to PLM fine-tuning, which is almost monotonically increasing with the training set size, adapter-tuning and prefix-tuning have fluctuations. From preliminary experiments, we find that adapter-tuning and prefix-tuning are more sensitive to learning rate than PLM fine-tuning. Fixing two separate learning rates for few shot and non-few shot experiments for all languages, a simplification of the normal training process to find an optimal learning rate for each training set size and each language on the development set, might cause the unstable performance of the tuning methods that are more sensitive to learning rate. (4) All above observations are roughly true for every language and are especially clear in the average plot. 

\paragraph{Discussion \& Takeaways} The possible explanation of the different behavior of adapter-tuning and prefix-tuning is their structural discrepancy, in which adapter-tuning adds parameters between two transformer layers, while prefix-tuning uses these parameters to generate prefixes and appends the prefixes at the front of each transformer layer. Similar to lengthening the input, prefix-tuning does not touch the PLM’s architecture, preserving the knowledge in PLM. Thus, by utilizing the PLM better, prefix-tuning has better performance at few shots when there are not enough samples to learn new patterns. However, when the training set size becomes larger, more flexible structures are needed to learn from these samples, leading to the better performance of adapter-tuning. The possible reason why both prefix-tuning and adapter-tuning outperform PLM fine-tuning at the left side is that prefix-tuning and adapter have only 8\% parameters to tune, which avoid overfitting when the training samples are not enough.

From this experiment, we can see that prefix-tuning has both advantages of utilizing the knowledge of PLM better and fewer parameters to tune, while adapter-tuning only benefits from the latter. This reminds us that while designing a prompt~\cite{liu2021pre}, one important thing is to keep the PLM’s architecture and make the prompt as natural as possible to better extract knowledge from PLM. 

\begin{table*}[!htp]\centering
\scriptsize
\begin{tabular}{lrrrrrrrrrrrrrr}\toprule
\multirow{3}{*}{Language} &\multirow{3}{*}{Script} &\multicolumn{6}{c}{mt5-base} &\multicolumn{6}{c}{mt5-base34} \\\cmidrule(lr){3-8} \cmidrule(lr){9-14}
& &\multicolumn{3}{c}{prefix-tuing} &\multicolumn{3}{c}{adapter-tuing} &\multicolumn{3}{c}{prefix-tuing} &\multicolumn{3}{c}{adapter-tuing} \\\cmidrule(lr){3-5} \cmidrule(lr){6-8}\cmidrule(lr){9-11} \cmidrule(lr){12-14}
& &R1 &R2 &RL &R1 &R2 &RL &R1 &R2 &RL &R1 &R2 &RL \\\midrule
amharic &Ge'ez &15.33 &5.42 &13.8 &16.58 &5.88 &14.88 &16.97 &5.88 &15.12 &17.85 &6.2 &16.01 \\
azerbaijani &Cyrillic &15.72 &6.3 &14.47 &17.81 &7.34 &16.13 &11.9 &3.4 &10.93 &12.16 &3.37 &10.96 \\
bengali &Brahmic &23.76 &9.11 &20.66 &25.99 &10.07 &22.21 &0 &0 &0 &0 &0 &0 \\
burmese &Brahmic &12.74 &3.46 &11.51 &14.07 &4.09 &12.54 &1.57 &0.38 &1.51 &1.14 &0.22 &1.08 \\
igbo &Latin &23.27 &5.25 &17.36 &25.22 &7.08 &19.38 &28.1 &7.98 &21.19 &28.33 &7.93 &21.72 \\
japanese &Kan,Hi,Kat &41.23 &18.89 &32.42 &45.6 &21.87 &35.02 &11.39 &2.82 &9.05 &11.84 &3.15 &9.12 \\
scottish\_gaelic &Latin &24.05 &7.73 &19.45 &20.42 &4.31 &17.02 &24.36 &7.16 &18.87 &26.37 &8.09 &20.22 \\
spanish &Latin &28.09 &9.05 &21.14 &29.28 &10.3 &22.23 &29.75 &9.48 &22.18 &30.26 &9.83 &22.42 \\
tamil &Brahmic &16.45 &6.58 &15 &19.81 &8.83 &18.12 &0.42 &0.02 &0.42 &0.44 &0.02 &0.43 \\
ukrainian &Cyrillic &18.73 &7.07 &16.43 &21.84 &8.92 &19.08 &16.46 &4.41 &13.99 &16.96 &4.61 &14.32 \\
urdu &Arabic &35.05 &14.5 &28.76 &38.81 &17.69 &32.08 &20.19 &4.48 &15.66 &19.57 &4.58 &15.08 \\
\bottomrule
\end{tabular}
\caption{R1, R2, and RL scores of prefix-tuning, adapter-tuning of mt5-base and mt5-base34 for 11 languages. ``Kan, Hi, Kat'' is the abbreviation of Kanji, Hiragana, and Katakana.
}
\label{tab: mt5-base34}
\end{table*}

\subsection{Exp-II: Effect of Supervised Transfer}   \label{exp2}
When designing models in multi-lingual scenarios, one crucial question is how to make modules of different languages communicate efficiently so that shared knowledge can be fully utilized.
There are two ways to do this: (1) by transferring: first fine-tune PLMs on multiple languages except the one we concerned about (a.k.a target language), and then adapt the fine-tuned multi-lingual model to the target language; (2) by multitask learning: jointly train all languages together. In this experiment, we study method 1 and try to answer the Q2: will supervised transfer be helpful for parameter-efficient tuning.  In Exp.\ref{exp3}, we study method 2.  


\subsubsection{Experiment Details}
\paragraph{Settings and Hyper-parameters}
We first divide \texttt{XL-Sum} dataset into two parts, one of which including 34 languages (75\% of total languages) is used to fine-tune PLM jointly on multiple languages to obtain a single multi-lingual model, while another part including 11 languages (25\% of total languages) is used to investigate the fine-tuned PLM's ability to generalize to new languages. The 11 left-out languages that do not participate in fine-tuning are chosen according the principle that they are from different language family and have different training set size.\footnote{Concretely, 7 languages (Amharic, Azerbaijani, Bengali, Burmese, Igbo, Japanese, Scottish Gaelic, Spanish, Tami) are low-resource ($<15,000$ training samples), 2 languages (Spanish, Tamil) are medium-resource ($15,000\sim 40,000$) and 2 languages (Ukrainian, Urdu) are high-resource ($>40,000$).} The hyper parameters used to fine-tune PLM is the same as the Multilingual training of \texttt{XL-Sum} \citep{hasan-etal-2021-xl}.\\
We refer to mt5-base as the original PLM without fine-tuning and mt5-base34 as the fine-tuned version on 34 languages.
We then performed prefix-tuning and adapter-tuning on mt5-base and mt5-base34 for 11 left-out languages. Hyperparameters used for these tuning methods remain the same as those in non-few shot experiments of Sec.(\ref{exp1}).

\subsubsection{Results and Analysis}
\paragraph{Results}

Table.\ref{tab: mt5-base34} illustrates the performance of multi-lingual models mt5-base and mt5-base34 to adapt to 11 new languages by prefix-tuning and adapter-tuning. The main observations in Table.\ref{tab: mt5-base34} are as follows:\\
(1) For four languages, Amharic, Igbo, Scottish Gaelic, and Spanish, mt5-base34 will bring gains against their counterparts by 0.3 to 6.0 R1 score.\\
(2) Fine-tuning mt5-base on 34 languages of \texttt{XL-Sum} dataset jeopardizes the performance of seven languages, among which Bengali, Burmese, Tamil's R1 score becomes near zero.\\
(3) Three of the four languages with performance improvement adapted from fine-tuned PLM are of the Latin script, while all three languages with dramatic performance drop down are of the Brahmic script, indicating the important role script plays to determine whether supervised transfer is helpful for parameter-efficient tuning. \\
\paragraph{Discussion \& Takeaways}
Although intuitively new languages will benefit from fine-tuning the PLM on the \texttt{XL-Sum} dataset, the practical results show that not all languages obtain improvements. Transfer learning in such a way might cause catastrophic forgetting of the previously acquired knowledge in PLM \cite{mccloskey1989catastrophic,santoro2016one}. If there are not enough training samples of a certain script during PLM fine-tuning, the PLM might lose the ability generalizing to languages of this script by parameter-efficient tuning methods and freezing PLM. This indicates that \textit{the effectiveness of parameter-efficient tuning methods under multi-lingual scenarios is highly dependent on the multi-lingual model we use and under some situations, parameter-efficient tuning might lose their adaptivity}. We leave how to alleviate this problem for future work.

\subsection{Exp-III: Joint Multi-Lingual Training}  \label{exp3}
In this experiment, we study joint multi-lingual training to see if different languages can benefit from each other, and also how adding private parameters for each language influences the performance of multi-lingual training (Q3).
\begin{figure*}[t]
\centering
\includegraphics[width=14cm]{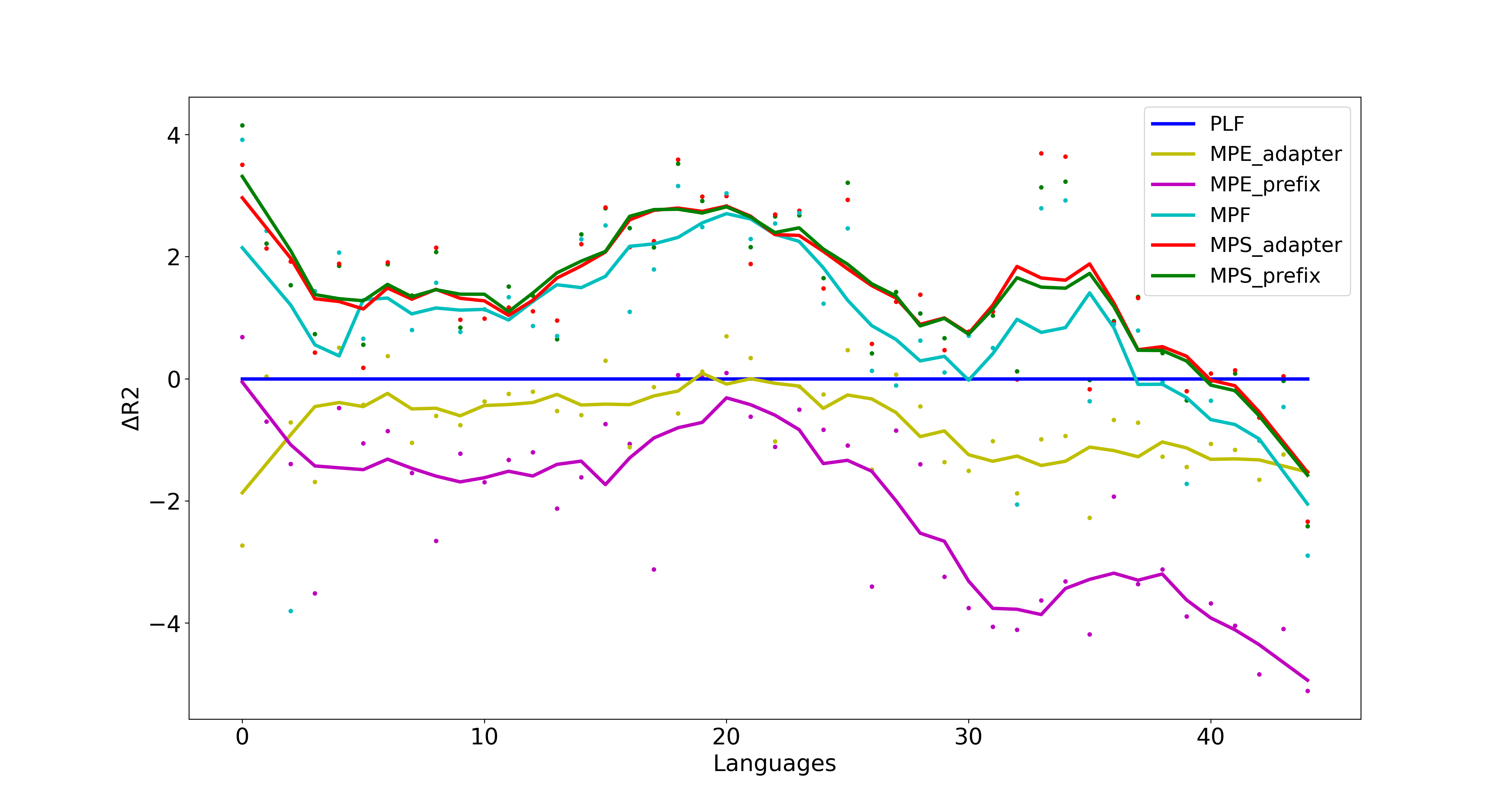}
\caption{Trend-lines depicting performance improvement. X-axis is the languages, which are arranged in increasing order of available training data from left to right. Y-axis depicts the R2 score relative to the singular language PLM fine-tuning baseline. } 
\label{single_vs_multi}
\end{figure*}

\subsubsection{Experiment Details}
With respect to Fig.\ref{fig:intro}, we have 6 different settings to compare single language training with multi-lingual training. 

\noindent\textbf{PLM Fine-tuning (PLF)}: mt5-base is fine-tuned on all languages to obtain separate models for each language as the baseline. 
\noindent\textbf{Multi-lingual Parameter-efficient (Adapter/Prefix) Tuning (MPE\_adapter/MPE\_prefix):} We add adapters/prefixes (with parameters = 8\% parameters of the LM) for each language and tune adapters/prefixes separately for each language while freezing mt5-base which is shared by all languages.
\noindent\textbf{Multi-lingual PLM Fine-tuning (MPF):} A single model is trained with training samples from multiple languages. The training strategy proposed by \citep{lample2019cross} to use a smoothing factor (alpha) of 0.5 to balance the sampling rate of low-resource languages and high-resource languages is followed by every multilingual setting.\footnote{We use the results of \texttt{XL-Sum}\citep{hasan-etal-2021-xl}. In order to have a fair comparison and remove the influence of different rouge packages, we use their model-generated outputs on test set to calculate the rouge score, instead of using their reported rouge score directly.}
\noindent\textbf{Multi-lingual Private-shared (Adapter/Prefix) Tuning (MPS\_adapter/MPS\_prefix):} Private adapters/prefixes (with parameters = 2\% parameters of LM) for each language (in total 45 languages in \texttt{XL-Sum} dataset $\times$ 2\% parameters for each languages = 90\% additional parameters) are added to a single LM. The LM is tuned jointly for multiple languages, while the adapters/prefixes are tuned separately for each language.%
\footnote{One thing to notice is that during training, in each iteration we sample from one language with a certain probability. The LM shared by all languages is tuned every iteration, while the private parameters for each language are tuned whenever the language is sampled.}

\subsubsection{Results and Analysis}
The summarization performance on different languages with different settings is plotted in Figure \ref{single_vs_multi}. With combinations of different experiment settings, we have the following results:\\
\textbf{PLF v.s. MPE\_prefix v.s. MPE\_adapter:} PLF outperforms MPE\_prefix and MPE\_adapter overall, with a gap larger for more available training samples. MPE\_adapter outperforms MPE\_prefix for almost every language, except a few languages with few samples. This conforms to the result in Sec.\ref{exp1} that fine-tuning has the advantage with large training set size, while prefix-tuning has the advantage with small training set size. One thing to notice is that adapter-tuning has comparable or even higher performance when the training set size is smaller than 10k, consistent with the result in Sec.\ref{exp1} and adapter’s sensitivity to training set size is not as high as prefix-tuning. The latter is reflected in that the adapter’s performance does not drop down dramatically as the training set size increases and keeps within -2 R2 of the baseline for almost all languages, which is even true for English with the highest training set size of 300,000. This means the parameter efficient adapter is a reasonable substitute of PLM fine-tuning regardless of the training set size.\\

\noindent \textbf{MPF v.s. PLF:} Multi-lingual model MPF significantly outperforms the baseline PLF in the low- and medium-resource languages with the gain decreasing as the training set size becomes larger. This is expected because low- and medium-resource languages can benefit from joint training by positive transfer between sister languages \cite{lample2019cross}. A deterioration is also observed in the high-resource languages. However, the losses are relatively minor; the multilingual model is within a -1 R2 drop for 6 high-resource languages and a -3 R2 drop for English. This indicates that by training a single multilingual model, the low- and medium-resource languages have been significantly improved without too much sacrifice in the high-resource languages. Similar to the low-resource experiment of \citep{hasan-etal-2021-xl}, our result is stronger than theirs, which only selects 5 low-resource languages to fine-tune individual LM.\\

\noindent \textbf{MPF v.s. MPS\_adapter v.s. MPS\_prefix:} By adding language-specific parameters in the multi-lingual scenario, compared to MPF, MPS\_adapter and MPS\_prefix have performance improvements R1:0.73, R2:0.46, RL:0.45 and R1:0.67, R2:0.47, RL:0.46 respectively for 45 languages on average. From each language performance, we can see that all high-resource languages have performance improvements at the cost of jeopardizing the performance of a few low-resource languages a little. This indicates that sharing LM as well as adding private language-specific parameters will maintain the jointly multi-lingual training’s advantage of sharing information among languages, while reducing the harm of sharing all parameters to high-resource languages due to the limit model capacity. \\
Besides, the two ways to add additional parameters: private adapter, private prefix for each language have roughly the same overall performance on the whole dataset and the same trend lines depicted in Fig.\ref{single_vs_multi}, despite their differences we have discussed in Sec.\ref{exp1}. The possible explanation is that the disadvantage of prefix-tuning lacking the flexibility to modify freeze LM addressed in Sec.\ref{exp1}, is alleviated or removed by tuning shared LM. Both prefix and adapter’s advantage comes from adding private parameters, so they have similar behavior. 

\section{Related Work}
\subsection{Multilingual Tasks}
With rapid development of pre-trained LMs, multilingual LMs have emerged to leverage the power of pre-training on a large number of languages, exemplified by 
\textbf{mBERT} \citep{BERT}, \textbf{XLM-R} \citep{conneau-etal-2020-unsupervised}, \textbf{XLM-R} \citep{conneau-etal-2020-unsupervised}, which adopt masked language model paradigm, and \textbf{mBART} \citep{liu-etal-2020-multilingual-denoising}, \textbf{mT5} \citep{xue-etal-2021-mt5}, which utilize a  sequence-to-sequence framework.
However, a few works have focused on multilingual summarization given the lack of benchmark datasets for other languages except English. 
\citep{multiling2015} benchmarked summarization systems over 40 languages, with limitation of dataset scale having 10k samples in total. \citep{MLSUM} released the multilingual summarization dataset spanning 5 languages with 1.5M article-summary pairs. \citep{cao2020multisumm} created a new dataset for two languages with 400k samples. \citep{hasan-etal-2021-xl} introduced \texttt{XL-Sum} spanning 45 languages containing 1.1M article-summary pairs. More recently, \citep{varab2021massivesumm} released MassiveSumm containing 28.8 million articles across 92 languages.

\subsection{Parameter Efficient Tuning}
Parameter-efficient tuning methods only tune a small number parameters to achieve comparable results.
It can be roughly divided into two categories, methods without additional parameters and methods with additional parameters. The former tune part of the pre-trained LM. \citep{lee2019elsa} fine-tunes a few of the final layers, while \citep{min2021noisy} only fine-tunes the bias terms of the LM. 
The latter introduces extra parameters while fixing the pre-trained LM. Popular methods  include \textbf{adapter-tuning} \citep{houlsby2019parameter,bapna2019simple,pfeiffer2020adapterhub} 
\textbf{prefix-tuning} \citep{li-liang-2021-prefix} 
\textbf{prompt-tuning} \citep{promptTuning},
and others \citep{mao2021unipelt,hu2021lora,diff-pruning}. Among these works, a comprehensive discussion in the context of multilingual summarization is relatively missing.

\section{Discussion}
In this paper, we investigate the applicable scope of different families of tuning strategies for multilingual learning.
We specifically ask three research questions, and by extensive experiments on summarization datasets with 45 languages we obtain diverse observations which, hopefully, would provide a useful instruction for future designing of multilingual tuning strategies.
\section{Limitations}
One limitation of our work is that we only conduct experiments on one summarization dataset, and more other NLP tasks could be explored as future work. Another limitation is that we only use ROUGE as our evaluation metric. More evaluation metrics, like BERTScore \citep{zhang2019bertscore}, COMET \citep{rei2020comet} or manual evaluation could be used. 




\bibliography{anthology,acl2020}
\bibliographystyle{acl_natbib}

\newpage 
\appendix
\section{Supplementary Material}
\subsection{Hyper-parameters}
\begin{table*}[!htp]\centering
\scriptsize
\begin{tabular}{lrrrrrrrrr}\toprule
&learning rate &epoch &batch size &grad acc &prefix length &inner dim &beam size &length penalty \\\midrule
prefix-tuning & & & & & & & & \\
few shot setting &3.00e-4 &20 &2 &1 &20 &800 &4 &0.6 \\
not few shot setting, low-resource &2.00e-4 &15 &8 &1 &200 &800 &4 &0.6 \\
not few shot setting, not low-resource &2.00e-4 &15 &16 &4 &200 &800 &4 &0.6 \\\cmidrule(lr){1-9}
adapter-tuning & & & & & & & & \\
few shot setting &1.00e-3 &20 &2 &1 &- &1200 &4 &0.6 \\
not few shot setting, low-resource &1.00e-3 &15 &8 &1 &- &1200 &4 &0.6 \\
not few shot setting, not low-resource &1.00e-3 &15 &16 &4 &- &1200 &4 &0.6 \\\cmidrule(lr){1-9}
PLM fine-tuning & & & & & & & & \\
few shot setting &5.00e-4 &20 &2 &1 &- &- &4 &0.6 \\
not few shot setting, low-resource &5.00e-4 &15 &8 &1 &- &- &4 &0.6 \\
not few shot setting, not low-resource &5.00e-4 &15 &16 &4 &- &- &4 &0.6 \\
\bottomrule
\end{tabular}
\caption{Hyper-parameter settings for Sec.\ref{exp1}. Grad acc is the abbreviation of gradient accumulation size, inner dim is the abbreviation of inner dimension.}
\label{tab: hp1}
\end{table*}

In Table\ref{tab: hp1}, we report the hyper-parameters used in Sec.\ref{exp1}.

\subsection{Additional Results}
Table\ref{tab: single-vs-multi} supplements Fig.\ref{single_vs_multi} in Sec.\ref{exp3}.

\label{sec:appendix}
\begin{table*}[!htp]\centering
\scriptsize
\begin{tabular}{lrrrrrrr}\toprule
language &PLF &MPE\_adapter &MPE\_prefix &MPF &MPS\_adapter &MPS\_prefix \\\midrule
amharic &17.46/6.64/16 &16.58/5.88/14.88 &15.33/5.42/13.8 &20.08/7.41/18.05 &20.6/7.61/18.5 &20.33/7.48/18.34 \\
arabic &34.82/15.13/29.22 &32.29/12.85/26.43 &29.02/10.94/24.14 &34.89/14.76/29.15 &35.16/14.96/29.27 &35.36/15.11/29.49 \\
azerbaijani &16.93/7.04/15.36 &17.81/7.34/16.13 &15.72/6.3/14.47 &21.4/9.55/19.37 &21.77/9.85/19.62 &21.58/9.83/19.76 \\
bengali &25.9/9.73/22.12 &25.99/10.07/22.21 &23.76/9.11/20.66 &29.46/12.02/25.1 &29.11/11.61/24.64 &29.57/11.88/24.89 \\
burmese &14.35/4.51/13.08 &14.07/4.09/12.54 &12.74/3.46/11.51 &16.17/5.17/14.42 &15.84/4.7/14.08 &16.12/5.07/14.39 \\
chinese simplified &40.87/25.97/34.09 &39.81/24.98/33 &36.3/22.34/30.22 &43.8/28.76/36.9 &44.95/29.66/37.79 &44.37/29.1/37.18 \\
chinese traditional &40.04/25.11/33.31 &39.16/24.17/32.3 &35.99/21.79/29.7 &43.21/28.03/36.17 &44.26/28.75/37.01 &43.83/28.33/36.66 \\
english &40.67/18.04/32.72 &38.96/16.52/31.06 &34.6/12.93/27.38 &37.61/15.15/29.88 &38.29/15.71/30.41 &38.29/15.63/30.39 \\
french &32.02/13.47/25.29 &31.85/13.34/25.13 &30.63/12.97/24.49 &35.33/16.19/28.2 &36.06/16.22/28.4 &35.69/16.15/28.16 \\
gujarati &19.38/6.49/17.6 &19.31/6.23/17.62 &18.03/5.65/16.45 &21.96/7.72/19.9 &22.45/7.97/20.21 &22.38/8.14/20.31 \\
hausa &36.16/15.43/29.13 &35.75/14.84/28.21 &34.08/13.82/27.07 &39.41/17.72/31.64 &39.75/17.64/31.85 &39.63/17.8/31.96 \\
hindi &38.64/17.33/32.38 &37.57/16.09/31.1 &34.32/13.23/28.23 &38.57/16.87/32.03 &39.18/17.37/32.49 &39.02/17.3/32.35 \\
igbo &27.16/8.76/21.37 &25.22/7.08/19.38 &23.27/5.25/17.36 &31.64/10.2/24.51 &30.17/9.2/22.98 &30.11/9.5/23.1 \\
indonesian &35.69/16.23/29.92 &34.86/15.51/28.85 &30.96/12.87/25.72 &37.01/17.02/30.75 &37.83/17.55/31.37 &37.69/17.57/31.46 \\
japanese &45.86/22.01/35.39 &45.6/21.87/35.02 &41.23/18.89/32.42 &48.08/23.8/37.32 &48.65/24.26/37.33 &48.41/24.16/37.39 \\
kirundi &28.97/12.84/23.6 &28.94/12.23/23.06 &26.18/10.18/20.51 &32/14.41/25.82 &32.65/14.98/26.22 &32.65/14.91/26.26 \\
korean &19.04/9.42/18.06 &19.92/9.93/18.59 &18.07/8.95/17.08 &23.7/11.49/22.34 &23.57/11.31/21.78 &23.11/11.27/21.6 \\
kyrgyz &13.66/5.58/12.43 &13.89/5.62/12.69 &11.77/4.88/10.87 &18.34/8.01/16.5 &18.4/7.72/16.0 &18.3/7.8/16.11 \\
marathi &20.21/8.94/18.31 &20.03/8.49/18.19 &18.3/7.55/16.83 &22.06/9.57/20.01 &23.13/10.32/20.8 &22.82/10.02/20.54 \\
nepali &23.16/9.06/21.14 &23.63/8.69/21.51 &20.91/7.37/19.25 &26.57/10.2/24.22 &26.5/10.05/24.06 &26.59/10.2/24.21 \\
oromo &16.55/5.35/14.61 &16.2/5.14/14.13 &13.7/4.14/12 &18.74/6.21/16.19 &19.68/6.45/16.91 &19.49/6.71/16.88 \\
pashto &37.69/15.38/31.19 &36.74/14.02/29.85 &33.85/12.14/27.88 &38.28/15.49/31.77 &38.85/15.85/32.05 &38.92/16.05/32.16 \\
persian &37.08/16.78/30.44 &35.72/15.34/28.81 &32.98/12.89/26.27 &35.71/15.06/29.1 &37.25/16.58/30.42 &37.32/16.43/30.3 \\
pidgin &34.55/12.67/27.03 &35.11/13.14/27.41 &32.85/11.57/25.79 &37.97/15.13/29.86 &38.56/15.6/30.14 &38.89/15.88/30.47 \\
portuguese &37.04/16.25/28.82 &36.3/15.18/27.7 &32.99/12.57/25.07 &37.15/15.89/28.53 &37.71/16.33/28.97 &37.59/16.21/28.91 \\
punjabi &26.18/9.61/21.95 &25.75/8.58/20.97 &25.94/8.49/21.43 &30.77/12.15/25.57 &31.14/12.3/25.33 &30.77/12.27/25.29 \\
russian &32.18/13.83/26.11 &30.56/12.67/24.73 &26.5/9.79/21.44 &32.21/13.64/26.16 &32.82/13.97/26.44 &32.61/13.92/26.39 \\
scottish gaelic &23.16/7.04/19.25 &20.42/4.31/17.02 &24.05/7.73/19.45 &29.01/10.96/22.87 &28.85/10.55/22.65 &30.08/11.2/23.83 \\
serbian cyrillic &18.64/4.83/15.51 &17.83/4.26/14.49 &18.52/4.89/15.58 &23.79/7.99/20.13 &24.5/8.42/20.73 &24.4/8.35/20.62 \\
serbian latin &17.13/4.19/14.24 &18.09/4.31/14.82 &18.04/4.21/14.9 &21.64/6.68/18.23 &22.91/7.18/19.26 &22.57/7.11/19.02 \\
sinhala &23.78/11.87/21.27 &23.29/11.16/19.81 &21.86/10.48/19.06 &21.51/8.07/18.9 &27.7/13.8/23.61 &27.43/13.41/23.7 \\
somali &29.08/10.21/22.45 &29.06/9.97/22.13 &28.09/8.88/21.45 &31.54/11.55/24.22 &32.1/11.38/24.42 &32.48/11.72/24.55 \\
spanish &30.05/10.98/23.13 &29.28/10.3/22.23 &28.09/9.05/21.14 &31.51/11.87/24.07 &31.63/11.9/24.11 &31.59/11.92/24.07 \\
swahili &33.82/14.82/27.78 &34.79/15.52/28.24 &33.43/14.91/27.19 &37.68/17.86/30.92 &38.24/17.82/31.16 &37.74/17.65/30.8 \\
tamil &22.47/10.33/20.58 &19.81/8.83/18.12 &16.45/6.58/15 &24.31/11.04/22.07 &24.58/11.11/22.3 &24.48/11.08/22.11 \\
telugu &16.66/5.84/15.1 &17.02/5.91/15.24 &15.14/4.99/13.77 &17.73/5.73/15.84 &20.09/7.1/17.78 &20.1/7.26/17.85 \\
thai &35.17/15.18/26.85 &36.08/14.06/25.7 &33.47/14.12/25.7 &36.43/16.28/28.22 &37.84/17.34/28.81 &37.99/17.65/29.07 \\
tigrinya &22.3/7.19/19.04 &21.37/6.14/17.89 &19.97/5.65/16.51 &25.26/7.99/21.1 &25.85/8.51/21.6 &25.48/8.55/21.78 \\
turkish &31.41/15.07/28 &30.53/14.04/26.97 &26.12/11/23.29 &32.92/15.57/29.28 &33.63/16.17/29.95 &33.58/16.1/29.81 \\
ukrainian &23.58/10.2/20.59 &21.84/8.92/19.08 &18.73/7.07/16.43 &23.99/10.14/20.92 &24.73/10.7/21.58 &24.75/10.62/21.57 \\
urdu &40.19/19.34/33.6 &38.81/17.69/32.08 &35.05/14.5/28.76 &39.49/18.33/32.83 &40.04/18.71/33.15 &40.12/18.71/33.21 \\
uzbek &13.48/4.84/12.32 &14.44/5.21/13.1 &11.77/3.98/10.88 &16.82/6.35/15.38 &17.45/6.74/15.73 &17.63/6.71/15.87 \\
vietnamese &32.53/16.45/25.94 &30.78/14.57/23.78 &27.33/12.34/21.27 &30.24/14.39/24.13 &33.62/16.43/26.46 &33.49/16.57/26.38 \\
welsh &31.58/11.46/25.6 &30.19/9.97/24.17 &27.92/8.06/22.24 &32.64/11.59/26.12 &33.09/12.03/26.41 &33.13/11.88/26.35 \\
yoruba &29.06/10.96/23.3 &29.94/10.43/23.31 &26.77/8.84/20.85 &31.62/11.66/25.06 &31.95/11.92/25.24 &31.71/11.61/24.91 \\
average &28.14/11.96/23.45 &27.58/11.23/22.66 &25.35/9.84/20.92 &30.23/12.93/25.11 &30.96/13.39/25.56 &30.89/13.4/25.57 \\
\bottomrule
\end{tabular}
\caption{R1/R2/RL of six models on 45 languages and the average score of all languages.}\label{tab: single-vs-multi}
\end{table*}

\end{document}